\documentclass[10pt,letterpaper]{article}

\usepackage{cogsci}

\cogscifinalcopy 

\usepackage{pslatex}
\usepackage{apacite}
\usepackage{float} 

\usepackage{cogsci}
\usepackage{pslatex}
\usepackage{apacite}
\usepackage{dblfloatfix}
\usepackage{subfig}

\usepackage{amsmath}
\usepackage{amsfonts}
\usepackage{amssymb}
\usepackage{graphicx}
\usepackage[round]{natbib}
\usepackage{color}

\usepackage{algorithm}
\usepackage[noend]{algpseudocode}

\usepackage{mathtools}

\title{Cognitive Models as Simulators:~The Case of Moral Decision-Making}
 
\author{{\large \bf Ardavan S. Nobandegani$^{1,2,5}$, Thomas~R.~Shultz$^{1,3}$, \& Irina Rish$^{4,5}$}\\
\{ardavan.salehinobandegani, thomas.shultz\}@mcgill.ca\\
\vspace*{-2pt}\{irina.rish\}@mila.quebec\vspace*{3pt}\\
\vspace*{-1pt}\small{$^{1}$Department of Psychology, McGill University}\\
\vspace*{-1pt}\small{$^{2}$Department of Electrical \& Computer Engineering, McGill University}\\
\vspace*{-1pt}\small{$^{3}$School of Computer Science, McGill University}\\
\vspace*{-1pt}\small{$^{4}$Department of Computer Science \& Operations Research, Universit\'e de Montr{\'e}al}\\
\vspace*{-1pt}\small{$^{5}$Mila - Quebec AI Institute}}

\begin{document}

\maketitle

\begin{abstract}
To achieve desirable performance, current AI systems often require huge amounts of training data.~This is especially problematic in domains where collecting data is both expensive and time-consuming,~e.g.,~where AI systems require having numerous interactions with humans, collecting feedback from them.~In this work, we substantiate the idea of \emph{cognitive models as simulators}, which is to have AI systems interact with, and collect feedback from, cognitive models instead of humans, thereby making their training process both less costly and faster.~Here,~we leverage this idea in the context of moral decision-making, by having reinforcement learning (RL) agents learn about fairness through interacting with a cognitive model of the Ultimatum Game (UG), a canonical task in behavioral and brain sciences for studying fairness. Interestingly, these RL agents learn to rationally adapt their behavior depending on the emotional state of their simulated UG responder. Our work suggests that using cognitive models as simulators of humans is an effective approach for training AI systems, presenting an important way for computational cognitive science to make contributions to AI.

\end{abstract}

\section{Introduction}
Recent years have witnessed artificial intelligence (AI) systems with remarkable abilities \citep[e.g.,][]{devlin2018bert,silver2016mastering,goyal2021self}, whose success critically depends on having access to huge amounts of training data. Examples include the famous Google BERT language model pre-trained on $800$M words from BooksCorpus and $2,500$M words from Wikipedia \citep{devlin2018bert}, the DeepMind AlphaGo system trained on over $30$M expert moves \citep{silver2016mastering}, the OpenAI GPT-3 model pre-trained on $300$ billion tokens \citep{brown2020language}, and the recent Facebook SEER image recognition model trained on one billion images from Instagram photos \citep{goyal2021self}.

Indeed, an influential subfield of AI, called reinforcement learning (RL), requires AI agents to learn by having interactions with their environment to collect feedback, in the form of rewards \citep{sutton2018reinforcement}. This is especially challenging in settings where the environment consists of human agents, resulting in these interactions being both expensive and time-consuming, thus exacerbating the training process. Could we instead use cognitive models, as a \emph{proxy} for humans, to address this issue?

In this work, we substantiate the idea of using cognitive models as {simulators}, which is to have AI systems interact with, and collect feedback from, cognitive models instead of humans, thereby making their training process both less costly and faster. Here, for the first time in the literature, we leverage this idea in the context of moral decision-making \citep{haidt2007new,lapsley2018moral}, by having RL agents learn about fairness through interacting with a cognitive model of the Ultimatum Game (UG), a well-established game in behavioral and brain sciences for studying fairness \citep[e.g.,][]{sanfey2009expectations,battigalli2015frustration,vavra2018expectations,sanfey2003neural,xiang2013computational,chang2013great}. Interestingly, these RL agents learn to rationally adapt their behavior depending on the emotional state of their UG Responder (see Sec.~1 for an explanation of how UG works). Our work suggests that using cognitive models as simulators of humans is an effective approach for training AI systems, presenting an important way for computational cognitive science to make contributions to the field of AI.

We begin by describing UG and presenting an overview of the relevant psychological findings on the role of emotions in UG (Sec.~2).~We then discuss in Sec.~3 a process model of UG Responder under a variety of emotional states (Lizzotte, Nobandegani, \& Shultz, 2021; Nobandegani, Destais, \& Shultz, 2020), and subsequently present our RL training results under various UG Responder's emotional states (Sec.~4).~We conclude by discussing the implications of our work for the fields of cognitive science and AI, and synergistic interactions between the two (Sec.~5).

\section{UG and the Role of Emotions in UG}
The Ultimatum Game (UG; \citealp{guth1982experimental}) is a canonical task for studying fairness, and has been extensively studied in psychology \citep[e.g.,][]{sanfey2009expectations,battigalli2015frustration,vavra2018expectations}, neuroscience \citep{sanfey2003neural,xiang2013computational,chang2013great}, philosophy \citep[][]{guala2008paradigmatic}, and behavioral economics \citep[e.g.,][]{guth1982experimental,thaler1988anomalies,camerer1995anomalies,fehr1999theory,sutter2003bargaining,camerer2006does}. UG has a simple design: Two players, Proposer and Responder, must agree on how to split a sum of money. Proposer makes an offer. If Responder accepts, the deal goes through; if Responder rejects, neither player gets anything. In both cases, the game is over. 

An extensive body of empirical work has established that UG Proposers predominantly respect fairness by offering about $50\%$ of the endowed amount, and that this split is almost invariably accepted by UG Responders \citep[see][]{camerer2011behavioral}.~Relatedly, UG Responders often reject offers below $30\%$,  presumably as retaliation for being treated unfairly \citep[][]{guth1982experimental,thaler1988anomalies,guth1990ultimatum,bolton1995anonymity,nowak2000fairness,camerer2006does}.

A growing body of experimental work has revealed that induced emotions strongly affect UG Responder's accept/reject behavior, with positive emotions increasing the chance of low offers being accepted \citep[e.g.,][]{riepl2016influences,andrade2009enduring}, and negative emotions decreasing the chance of low offers being accepted \citep[e.g.,][]{bonini2011pecunia,harle2010effects,liu2016negative,moretti2010disgust,vargas2019blocking}. Experimentally, these emotions are often induced by a movie clip or recall task.

\section{A Computational Model of UG Responder}
Recently, \citet{nobandegani2020UG} presented a process model of UG Responder, called \emph{sample-based expected utility} (SbEU). SbEU provides a unified account of several disparate empirical findings in UG (i.e., the effects of expectation, competition, and time pressure on UG Responder), and also explains the effect of a wide range of emotions on UG Responder (Lizzotte, Nobandegani, \& Shultz, \citeyear{nobandegani2021EmotionsinGames}).

Nobandegani et al.'s process-level account rests on two main assumptions.~First, UG Responder uses SbEU to estimate the expected-utility gap between their expectation and the offer, i.e., $\mathbb{E}[u(\text{offer})-u(\text{expectation})]$, where $u(\cdot)$ denotes Responder's utility function. If this estimate is positive --- indicating that the offer made is, on average, higher than Responder's expectation --- Responder accepts the offer; otherwise, Responder rejects the offer. This assumption is supported by substantial empirical evidence showing that Responder's expectation serves as a reference point for subjective valuation of offers \citep{sanfey2009expectations,battigalli2015frustration,vavra2018expectations,xiang2013computational,chang2013great}.

The second assumption is that negative emotions elevate loss-aversion while positive emotions lower loss-aversion \citep{nobandegani2021EmotionsinGames}.~Again, this assumptions is supported by mounting empirical evidence \citep[e.g.,][]{de2010amygdala,sokol2015interoceptive,sokol2009thinking} suggesting that emotions modulate loss-aversion --- the tendency to overweight losses as compared to gains (Kahneman \& Tverskey, \citeyear{Kahnemna1979}). 

Concretely, SbEU assumes that an agent estimates expected utility:
\begin{eqnarray}
\mathbb{E}[u(o)]=\int p(o)u(o)do,
\label{eq_exp_utility}
\end{eqnarray}
using self-normalized importance sampling (Nobandegani et al., 2018; Nobandegani \& Shultz, 2020b, 2020c), with its importance distribution $q^\ast$ aiming to optimally minimize mean-squared error (MSE): 
\begin{eqnarray}
\hat{E}=\dfrac{1}{\sum_{j=1}^s w_j} \sum_{i=1}^s w_i u(o_i), \quad \forall i:\ o_i\sim q^\ast,\ w_i=\dfrac{p(o_i)}{q^\ast(o_i)},
\label{eq_norm_is_estimator}
\end{eqnarray}
\begin{eqnarray}
q^\ast(o)\propto p(o)|u(o)| \sqrt{\dfrac{1+|u(o)|\sqrt{s}}{|u(o)|\sqrt{s}}}.
\label{eq_meta_rational_q}
\end{eqnarray}
MSE is a standard measure of estimation quality, widely used in decision theory and mathematical statistics (Poor, \citeyear{poor2013introduction}). In Eqs.~(\ref{eq_exp_utility}-\ref{eq_meta_rational_q}), $o$ denotes an outcome of a risky gamble, $p(o)$ the objective probability of outcome $o$, $u(o)$ the subjective utility of outcome $o$, $\hat{E}$ the importance-sampling estimate of expected utility given in Eq.~(\ref{eq_exp_utility}), $q^\ast$ the importance-sampling distribution, $o_i$ an outcome randomly sampled from $q^\ast$, and $s$ the number of samples drawn from $q^\ast$. 

SbEU has so far explained a broad range of empirical findings in human decision-making, e.g., the fourfold patterns of risk preferences in both outcome probability and outcome magnitude (Nobandegani et al., \citeyear{nobandegani2018meta}), risky decoy and violation of betweenness \citep{nobandegani2019BTW}, violation of stochastic dominance \citep*{nobandegani2022VoSD}, violation of cumulative independence \citep*{nobandegani2022VoCI}, the three contextual effects of similarity, attraction, and compromise (da Silva Castanheira, Nobandegani, Shultz, \& Otto, \citeyear{Castanheira2019value}), the Allais, St.~Petersburg, and Ellsberg paradoxes \citep{nobandegani2020resource,nobandegani2020bSTPP,nobandegani2021AllaisEllsberg}, cooperation in Prisoner's Dilemma \citep{nobandegani2019PD}, and human coordination behavior in coordination games \citep{nobandegani2020CG}.~Notably, SbEU is the first, and thus far the only, resource-rational process model that bridges between risky, value-based, and game-theoretic decision-making.

\section{Training RL Agents in UG}
\label{sec_rl_thompson_sampling}
In this section, we substantiate the idea of \emph{cognitive models as simulators} in the context of moral decision-making, by having RL agents learn about fairness through interacting with a cognitive model of UG Responder \citep{nobandegani2020UG}, as a proxy for human Responders, thereby making their training process both less costly and faster.

To train RL Proposers, we leverage the broad framework of multi-armed bandits in reinforcement learning \citep{katehakis1987multi,gittins1979bandit}, and adopt the well-known Thompson Sampling method \citep{thompson1933likelihood}. Specifically, we assume that RL Proposer should decide what percentage of the total money $T$ they are willing to offer to SbEU Responder. For ease of analysis, here we assume that RL Proposer chooses between a finite set of options: $\mathcal{A} = \{0, \frac{T}{10}, \frac{2T}{10}, \cdots, \frac{9T}{10}, T\}$.
\begin{algorithm}
\caption{{Thompson Sampling for UG Proposer}}\label{TS_proposer}
\begin{algorithmic}[1]
\Statex \textbf{Initialize}. $\forall a\in\mathcal{A}$: $S_{a} = 0$ and $F_{a} = 0$
\State \textbf{for} $i = 1,\ldots, N$
\State $\forall a\in\mathcal{A}$ compute:

$s_{a} = u(T - a)\beta_{a}$,\quad $\beta_{a}\sim$ Beta$(S_{a}+1, F_{a}+1)$ 
\State $a^\ast = \arg\max_{a}\,\,\, s_{a}$
\State Offer $a^\ast$ to SbEU Responder
\State \textbf{if} SbEU Responder accepts the offer \textbf{then}
\State \hspace*{20pt}$S_{a^\ast} = S_{a^\ast} + 1$
\State \textbf{else}
\State \hspace*{20pt}$F_{a^\ast} = F_{a^\ast} + 1$
\State \textbf{end if}
\State \textbf{end for}
\end{algorithmic}
\end{algorithm}

In reinforcement learning terminology, RL Proposer learns, through trial and error while striking a balance between exploration and exploitation, which option $a\in\mathcal{A}$ yields the highest mean reward. Here, we train RL Proposers using Thompson Sampling, a well-established method in the reinforcement learning literature enjoying strong optimality guarantees \citep{agrawal2012analysis,agrawal2013further}; see Algorithm 1.

Algorithm 1 can be described in simple terms as follows. At the start, i.e., prior to any learning, the number of times an offer $a\in\mathcal{A}$ is so far accepted, $S_a$ (S for success), and the number of times it is rejected, $F_a$ (F for failure), are both set to zero. In each trial (for a total of $N$ trials), an estimate of mean reward for each offer $a\in\mathcal{A}$ is computed by sampling from the corresponding distribution (Line 2), and the offer with the highest mean reward estimate $a^\ast$ (Line 3) is then chosen by Proposer to be offered to SbEU Responder (Line 4). If this offer is accepted by SbEU Responder, the $S_a$ parameter for that offer is incremented by one (Line 6); if rejected, the $F_a$ parameter for that offer is instead incremented by one. In Algorithm 1, $T$ is the total amount of money to be split between Proposer and Responder, $u(\cdot)$ is the subjective utility function of Responder, and Beta$(\cdot,\cdot)$ is the Beta distribution.

\begin{figure}[h!]
\centering
\includegraphics[trim = 178pt 8pt 145pt 10pt,clip,width=0.49\textwidth]{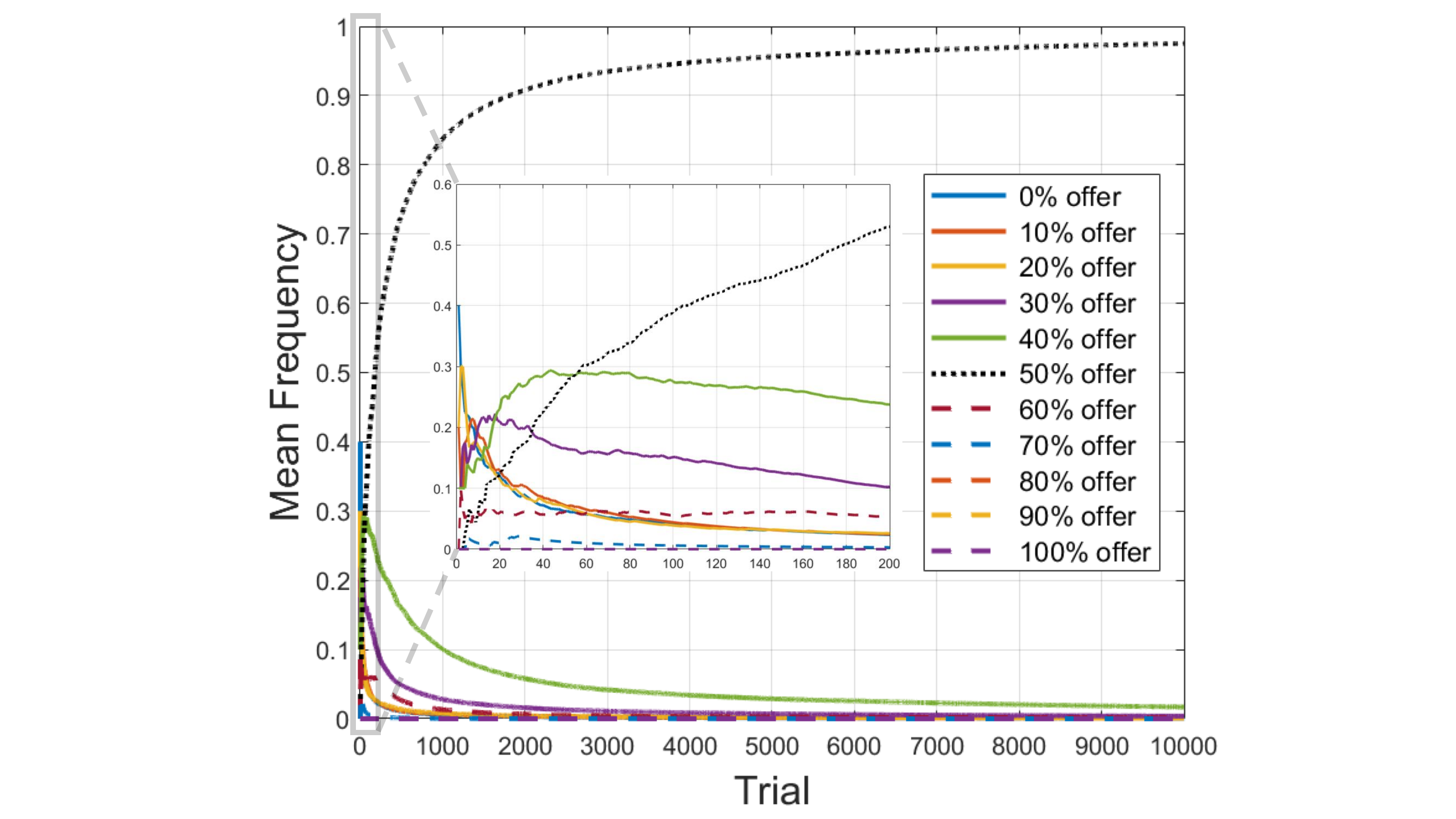}
\caption{\textbf{Mean frequency of RL Proposer's offers.} The $y$-axis indicates the mean frequency of each offer made by RL Proposer to SbEU Responder up to current trial ($x$-axis), averaged over 10 RL Proposers. SbEU Responder is in a neutral emotional state. As a visual aid, the dynamics for the first 200 trials are provided in a smaller plot, located at the center.}
\label{fig_neural}
\end{figure}

In Fig.~\ref{fig_neural}, we simulate 10 RL Proposers, and report the mean frequency of an offer being made to SbEU Responder over the past trials, for a total of $N=10,000$ trials. As can be seen, exercising a balance between exploration and exploitation, RL Proposers eventually arrive at the decision that they should be making a fair offer to SbEU Responder, i.e., to split the total sum $T$ equally between themselves and Responder. As such, RL Proposer's making fair offers can be seen as a rational, emergent behavior arising from having sufficient interactions with inequality-averse SbEU Responder --- as a proxy for inequality-averse humans.

\subsection{RL Proposer Meets Emotional Responder}
In this section, we bridge between the idea of \emph{cognitive models as simulators} and emotion research, by letting AI systems interact with a cognitive model of people experiencing various emotional states. Specifically, we pursue this idea in the context of UG, and have RL Proposers interact with SbEU Responders experiencing various emotional states.

A wealth of empirical research has revealed that the effect of emotions on human decision-making is both substantial and systematic \citep[for reviews see, e.g.,][]{phelps2014emotion,lerner2015emotion}. More specifically, in the context of UG, a growing body of empirical studies have shown that induced emotions strongly affect UG Responder's behavior, with positive emotions (e.g., happiness) increasing the chance of low offers being accepted \citep[e.g.,][]{riepl2016influences,andrade2009enduring}, and negative emotions (e.g., disgust, anger, and sadness) decreasing the chance of low offers being accepted \citep[e.g.,][]{bonini2011pecunia,harle2010effects,liu2016negative,moretti2010disgust,vargas2019blocking}. Hence, it would be rational for UG Proposer (from the perspective of maximizing their mean reward) to make larger offers to Responders experiencing negative emotions, and, conversely, to make smaller offers to Responders experiencing positive emotions.
\begin{figure}[h!]
\centering
\includegraphics[trim = 150pt 8pt 165pt 10pt,clip,width=0.48\textwidth]{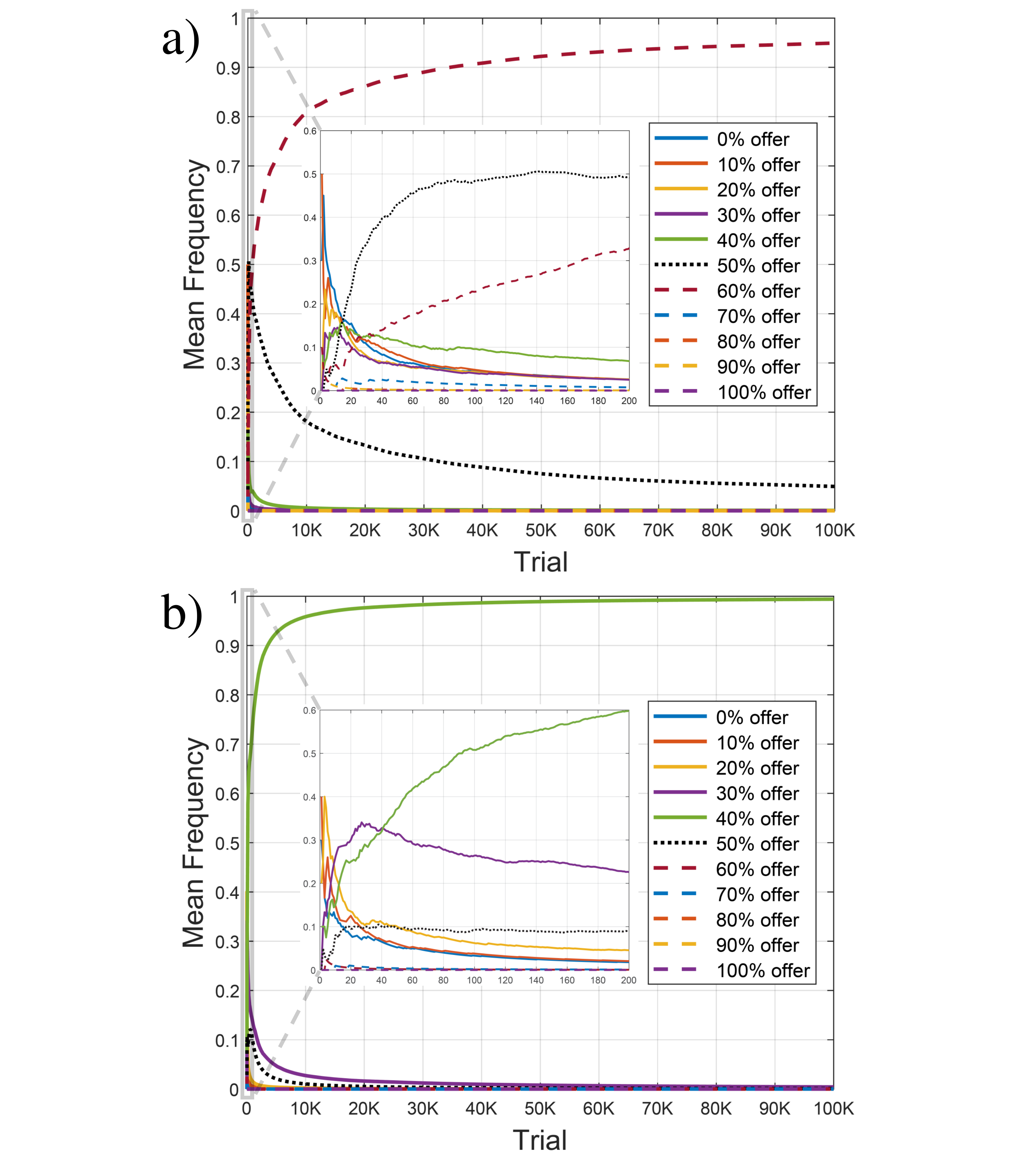}
\caption{\textbf{Mean frequency of RL Proposer's offers.} The $y$-axis indicates the mean frequency of each offer made by RL Proposer to SbEU Responder up to current trial ($x$-axis), averaged over 10 RL Proposers. In \textbf{(a)} SbEU Responder is under a negative emotional state, while in \textbf{(b)} SbEU Responder is under a positive emotional state. As a visual aid, in each subplot, the dynamics for the first 200 trials are provided in a smaller plot, located at the center.}
\label{fig_positive&negative}
\end{figure}

Interestingly, under the broad and empirically well-supported assumption that emotions modulate loss-aversion \citep[e.g.,][]{de2010amygdala,sokol2015interoceptive,sokol2009thinking}, Nobandegani et al.'s SbEU model explains the effect of a wide range of emotions on human UG Responder \citep{nobandegani2021EmotionsinGames}. Next, we train RL Proposers, using Thompson Sampling (see Algorithm \ref{TS_proposer}), to learn how to interact with SbEU Responders experiencing positive or negative emotional states.

In Fig.~\ref{fig_positive&negative}, we simulate 10 RL Proposers, and report the mean frequency of an offer being made to SbEU Responder over the past trials, for a total of $N=100,000$ trials. In Fig.~\ref{fig_positive&negative}(a), SbEU Responder is under a negative emotional state, and, in Fig.~\ref{fig_positive&negative}(b), SbEU Responder is under a positive emotional state. As can be seen, RL Proposers eventually arrive at the decision that they should be making a larger offer ($60\%$) when Responder is experiencing a negative emotional state (Fig.~\ref{fig_positive&negative}(a)), and, conversely, should be making a smaller offer ($40\%$) when Responder is experiencing a positive emotional state (Fig.~\ref{fig_positive&negative}(b)). As such, RL Proposers learn to adapt their strategy depending on Responder's emotional state.

\section{General Discussion}
To achieve desirable performance, current AI systems often require huge amounts of training data.~This is especially problematic in domains where collecting data is both expensive and time-consuming,~e.g.,~where AI systems require many interactions with humans, collecting feedback from them.~In this work, we substantiate the idea of \emph{cognitive models as simulators}, which is to have AI systems interact with, and collect feedback from, cognitive models as a proxy for humans, thereby decreasing both cost and time for the AI training process.

Here, for the first time in the literature, we leverage this idea in the context of moral decision-making \citep{haidt2007new,lapsley2018moral}, by having reinforcement learning (RL) agents learn about fairness through interacting with a cognitive model of the Ultimatum Game (UG), a canonical task for studying fairness in behavioral and brain sciences \citep[e.g.,][]{sanfey2009expectations,battigalli2015frustration,vavra2018expectations,sanfey2003neural,xiang2013computational,chang2013great}. As a cognitive model, we use {sample-based expected utility} (SbEU), a resource-rational process model explaining a wide range of empirical findings on UG Responders \citep{nobandegani2020UG,nobandegani2021EmotionsinGames}. As an AI system, we train RL Proposers using Thompson Sampling, a well-known method in the multi-armed bandits literature enjoying strong optimality guarantees \citep{agrawal2012analysis,agrawal2013further}.

Given the significant role that emotions play in human decision-making \citep[for reviews see, e.g.,][]{phelps2014emotion,lerner2015emotion}, we further link the idea of \emph{cognitive models as simulators} to emotion research, by having RL Proposers interact with SbEU Responders under various emotional states (i.e., neutral, negative, and positive). Interestingly, RL Proposers learn to rationally adapt their behavior depending on the emotional state of their SbEU Responder, making larger offers when Responder is more likely to reject low offers (due to experiencing negative emotions) and, conversely, making smaller offers when Responder is less likely to reject low offers (due to experiencing positive emotions).

Recent success stories in AI, e.g., AlphaGo and particularly self-play \citep{silver2016mastering,silver2017mastering}, clearly demonstrate the significant role that having access to a simulator of the environment would play in efficient training of AI systems. The idea of \emph{cognitive models as simulators} substantiated in our work is yet another step in the direction of leveraging simulators of the environment --- by using cognitive models as a proxy for people --- in the service of making the training of AI systems both faster and less costly. As such, the idea of \emph{cognitive models as simulators} presents an important way for computational cognitive science to contribute to AI.

Although here we presented the idea of \emph{cognitive models as simulators} as a way of making the training of AI systems more efficient, it could also be seen as a broad \emph{cognitive} framework for how people might be choosing their strategies in multi-agent environments by {mentalizing} about other agents. As such, the idea of \emph{cognitive models as simulators} could potentially serve as a broad {framework} for theorizing about, and mathematically identifying, mental processes by which people choose their strategies when interacting with other agents. Hence, this ``cognitive" reconceptualization of \emph{cognitive models as simulators} has potential to make contributions to computational cognitive science. 

Additionally, a strong reading of this cognitive reconceptualization takes the AI systems learning from interacting with mental models as a proposal for how people might be choosing their strategy in multi-agent environments, thus presenting an important way for AI to contribute to computational cognitive science.

From this perspective, the Thompson Sampling algorithm presented in Sec.~\ref{sec_rl_thompson_sampling} for training RL Proposers could serve as a {process-level proposal} for how human Proposers might be choosing their offer: by simulating UG Responder using a mental model of UG Responder and learning from mentally interacting with that model, here implemented by SbEU \citep{nobandegani2020UG}.~Nonetheless, human Proposers might be using a much simpler mental model of their human Responder as compared to SbEU, and would presumably start with much stronger prior beliefs (i.e., inductive biases) about the mean reward of each of their strategies --- instead of the uniformly distributed Beta$(1,1)$ prior used in Algorithm 1.~Future work should more extensively investigate this process-level proposal.

Also, this cognitive reconceptualization is consistent with substantial work on both people's intuitive psychology and human strategic decision-making \citep[e.g.,][]{jern2017people,jara2016naive,nagel1995unraveling,baker2009action,camerer2004cognitive}, broadly assuming that people have a mental model of other agents and use that model to both interpret other agents' behavior and decide how to behave when interacting with those agents.

Finally, as elaborated above, having good potential to contribute to both AI and computational cognitive science, the idea of \emph{cognitive models as simulators} is yet another step in the fruitful direction of having ever more synergistic interaction between the two disciplines. We see our work a step in this important direction.\\

\noindent \textbf{Acknowledgments}. This research was supported in part by an operating grant to TRS from the Natural Sciences and Engineering Research Council of Canada.

\setlength{\bibleftmargin}{.125in}
\setlength{\bibindent}{-\bibleftmargin}

\nocite{nobandegani2018meta}

\bibliographystyle{apacite}
\bibliography{arXiv_cogmod_as_simulators_ref}
\end{document}